\documentclass[10pt,twocolumn,letterpaper]{article}

\usepackage[pagenumbers]{wacv} 

\usepackage{graphicx}
\usepackage{amsmath}
\usepackage{amssymb}
\usepackage{booktabs}

\usepackage{tabto}
\usepackage{amsfonts}
\usepackage{algorithm}
\usepackage{algpseudocode}
\usepackage{multirow}

\usepackage[pagebackref,breaklinks,colorlinks]{hyperref}

\usepackage[capitalize]{cleveref}
\crefname{section}{Sec.}{Secs.}
\Crefname{section}{Section}{Sections}
\Crefname{table}{Table}{Tables}
\crefname{table}{Tab.}{Tabs.}

\def\wacvPaperID{1444} 
\def\confName{WACV}
\def\confYear{2024}

\begin{document}

\title{SSVOD: {S}emi-{S}upervised {V}ideo {O}bject {D}etection with Sparse Annotations}

\author{Tanvir Mahmud$^{1}$\thanks{Work done while interned at Bosch Research North America} \ \  Chun-Hao Liu$^{2}$ \ Burhaneddin Yaman$^{3}$ \ Diana Marculescu$^{1}$ \\
$^{1}$University of Texas at Austin \ \ \ $^{2}$Amazon Prime Video \\ $^{3}$Bosch Research North America\\
{\tt\small \{tanvirmahmud, dianam\}@utexas.edu}, \ \ \  {\tt\small chunhaol@amazon.com} \\ {\tt\small burhaneddin.yaman@us.bosch.com}
}

\maketitle
\thispagestyle{empty}

\begin{abstract}

Despite significant progress in semi-supervised learning for image object detection, several key issues are yet to be addressed for video object detection: (1) Achieving good performance for supervised video object detection greatly depends on the availability of annotated frames. (2) Despite having large inter-frame correlations in a video, collecting annotations for a large number of frames per video is expensive, time-consuming, and often redundant. (3) Existing semi-supervised techniques on static images can hardly exploit the temporal motion dynamics inherently present in videos. In this paper, we introduce SSVOD, an end-to-end semi-supervised video object detection framework that exploits motion dynamics of videos to utilize large-scale unlabeled frames with sparse annotations. To selectively assemble robust pseudo-labels across groups of frames, we introduce \textit{flow-warped predictions} from nearby frames for temporal-consistency estimation. In particular, we introduce cross-IoU and cross-divergence based selection methods over a set of estimated predictions to include robust pseudo-labels for bounding boxes and class labels, respectively. To strike a balance between confirmation bias and uncertainty noise in pseudo-labels, we propose confidence threshold based combination of hard and soft pseudo-labels. 
Our method achieves significant performance improvements over existing methods on ImageNet-VID, Epic-KITCHENS, and YouTube-VIS datasets. Code and pre-trained models will be released.
\end{abstract}

\begin{figure}[t]
  \centering
   \includegraphics[width=0.9\linewidth]{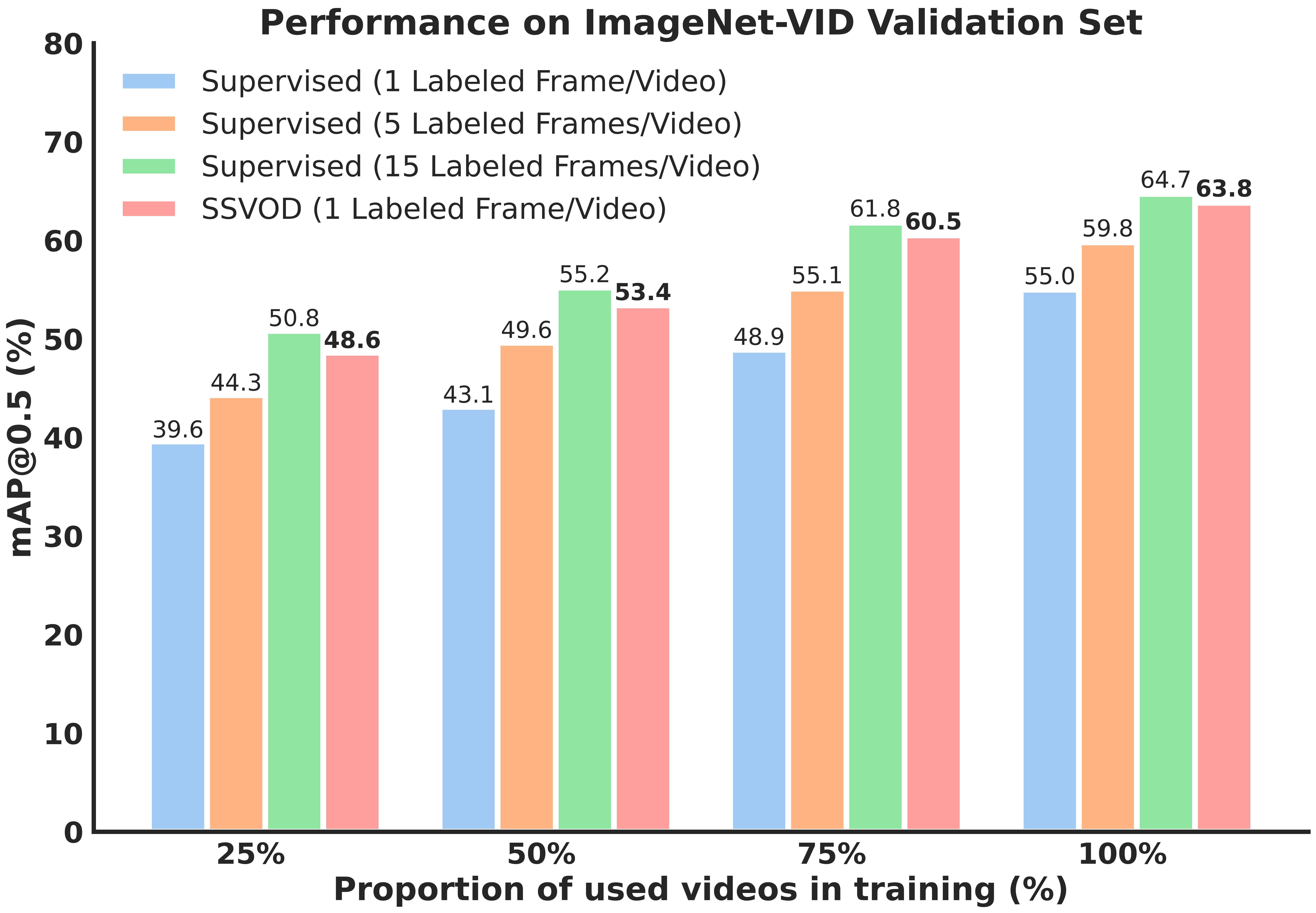}
   \caption{Supervised performance greatly depends on the availability of annotated frames per video. Our proposed SSVOD largely reduces the performance gap between sparse annotations (1 frame/video) and dense annotations (15 frames/video).}
   \label{f1}
\end{figure}

\section{Introduction}
Human annotations are expensive, time-consuming, and hard to collect for large-scale datasets~\cite{weakly}. In this regard, semi-supervised learning has great potential to utilize large-scale unlabeled data with limited annotations~\cite{survey, mixmatch, s4l}. In recent years, researchers have shown significant progress in semi-supervised learning (SSL) on many applications, such as image classification~\cite{mixmatch, pmlr-v162-wang22s}, semantic segmentation~\cite{8954327,10.1007/978-3-030-58545-7_40}, and object detection~\cite{sv2, instant}. However, prior work for semi-supervised object detection has mostly focused on image-based methods which opens up several key issues to be addressed particularly for videos. In this paper, we aim to fill this gap by redesigning the SSL architecture for video object detection.

\begin{figure}[t]
  \centering
   \includegraphics[width=0.8\linewidth]{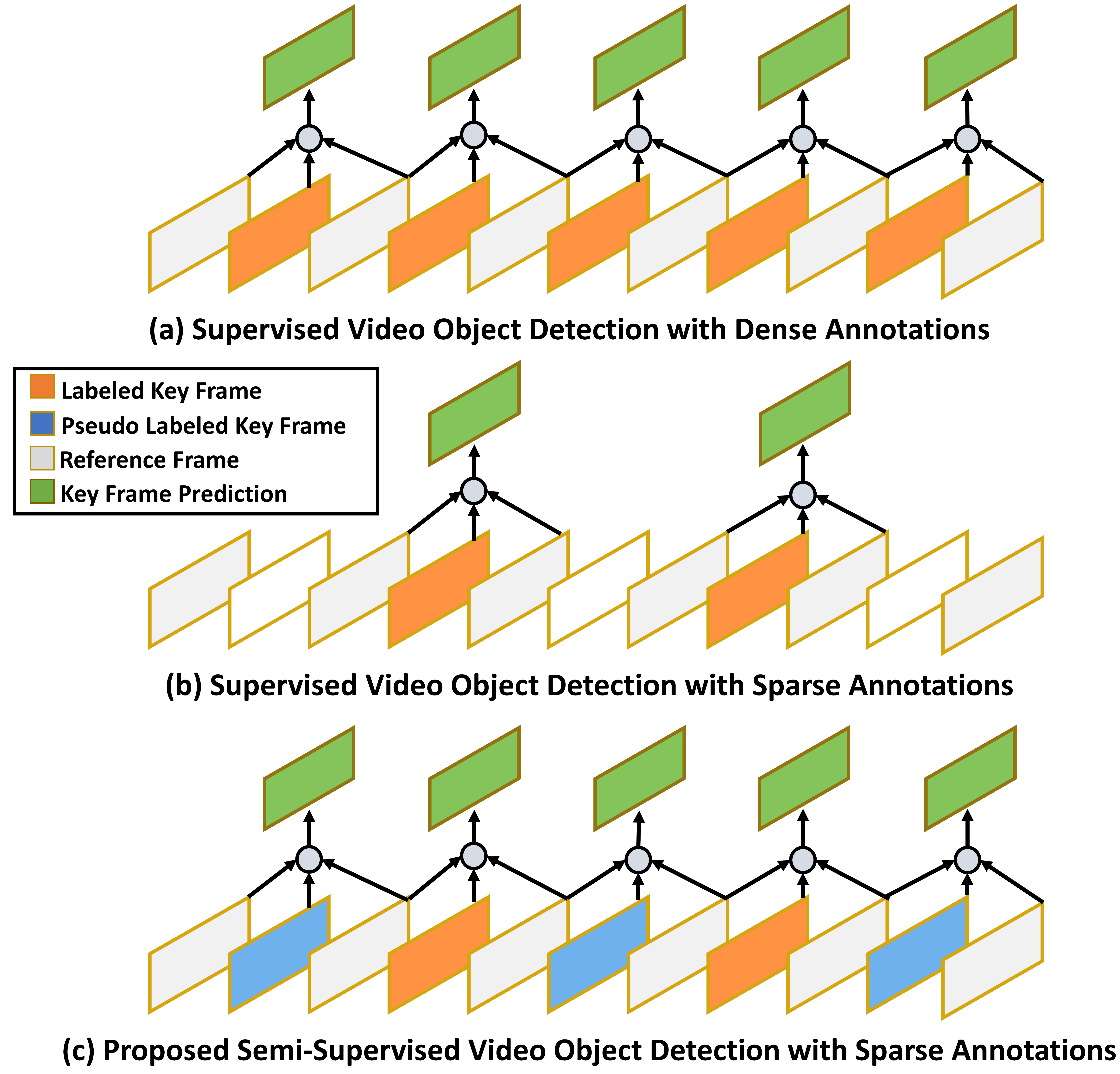}
   \caption{
   (a) Supervised video object detection requires \emph{dense annotations} to explore different time-steps of videos. (b) With \emph{sparse annotations}, the supervised method generates sub-optimal performance due to insufficient temporal exploration.
   (c) SSVOD leverages semi-supervised learning to \emph{estimate robust pseudo-labels} across different time-steps by exploiting available \emph{sparse annotations}, thereby reducing annotation burden.}
   \label{f3}
\end{figure}

\begin{figure*}[t]
  \centering
   \includegraphics[width=0.8\linewidth]{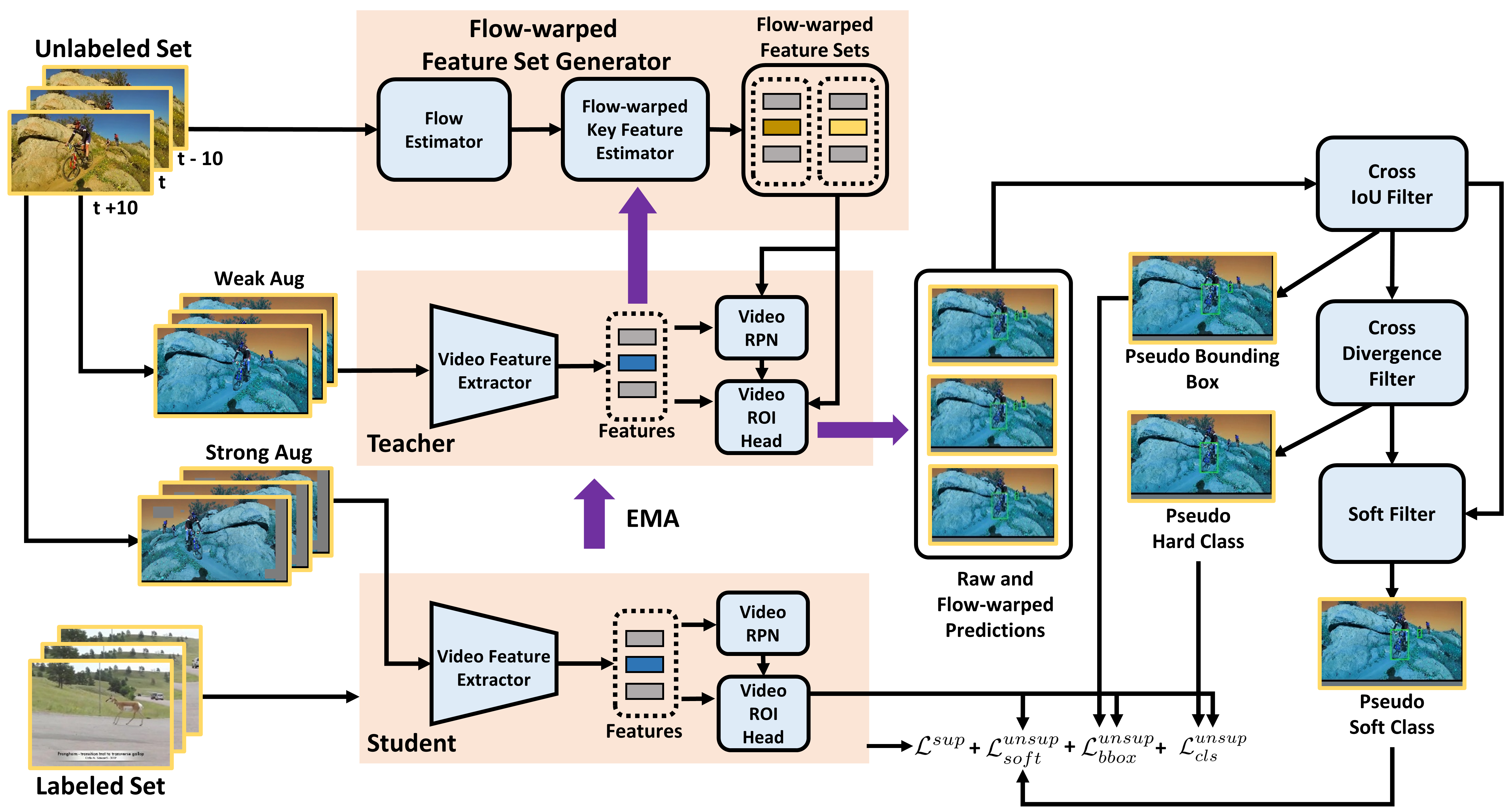}
   \caption{The overview of the semi-supervised video object detection (SSVOD). Each labeled/unlabeled set consists of one \textit{key} and several \textit{reference frames}. The teacher video detector operates on the unlabeled set to generate pseudo-labels. A flow-warped feature generator warps reference features using motion flow estimates to generate flow-warped feature sets. The \textit{key frame} predictions generated from the raw and flow-warped feature sets are processed with a three-stage filtering scheme to separate pseudo bounding boxes, hard class labels, and soft-labels. The student video detector is trained on unlabeled sets with pseudo-labels, as well as on labeled sets.}
   \label{f2}
\end{figure*}

Video comes with additional challenges compared to static images, particularly arising from motion deblurring, pose variations, and camera defocus under fast motion~\cite{wu2019selsa, fgfa}. 
Numerous approaches have been studied to exploit the rich temporal information  presented in videos to improve video object detection~\cite{tcnn, tubelet}. However, the performance of supervised video object detectors greatly depends on the availability of annotated frames per video. In case of extremely sparse annotations of one labeled frame per video, an average of 22\% supervised performance reduction is observed compared to the dense supervision of $15$ labeled frames per video (Fig.~\ref{f1}). For sparse annotations, standard supervised training strategies cannot explore different time-steps over the video, which results in sub-optimal performance (Fig. \textcolor{blue}{\ref{f3}b}). Despite having large inter-frame correlations in a video, collecting annotations for a large number of frames per video is expensive, time-consuming, and often redundant. Our primary motivation is to \textit{reduce the performance gap between sparse and dense annotations by fully exploiting the inherent temporal dynamics of videos}.

The advancement of semi-supervised image object detection has
introduced several key challenges in pseudo-label estimation for unlabeled images~\cite{soft, liu2021unbiased, unbiasedv2}. Wrong estimation of pseudo-labels for object classes and bounding boxes results in confirmation bias that diminishes the advantage of additional unlabeled data~\cite{confirmation}. Image-based techniques offer various pseudo-label selection and filtering techniques to systematically filter robust pseudo-labels for unlabeled images~\cite{soft, unbiasedv2}. However, such techniques customized for static images cannot utilize the rich temporal information presented in videos for searching reliable pseudo-labels across groups of frames.
Therefore, naive integration of image-based pseudo-label selection techniques into state-of-the-art (SOTA) video object detectors yield sub-optimal performance, thereby demanding more specialized solutions for videos (Table~\ref{t1}).

Moreover, existing semi-supervised learning approaches on videos~\cite{sv1, sv2} primarily focus on the post-processing of detected bounding boxes to estimate pseudo-labels, which are referred as \textit{box-level methods}~\cite{fgfa}. However, SOTA video object detectors mainly operate on the feature space of groups of frames to aggregate motion cues from surrounding \emph{reference frames} into the target \emph{key frame} prediction~\cite{soft, liu2021unbiased, unbiasedv2}. 
Hence, instead of operating on final image-level predictions of video detectors, it is necessary to exploit the temporal feature space of groups of frames to estimate most consistent pseudo-labels in the target \emph{key frame}.

In this paper, we introduce SSVOD, a semi-supervised learning framework that exploits the motion cues present in videos to greatly reduce annotation burden of SOTA video object detectors  (Fig. \textcolor{blue}{\ref{f3}c}). Inspired by semi-supervised image detectors, we propose a teacher-student framework in video object detection to train on sparsely labeled data. Instead of only operating on final image-level predictions of target \textit{key frames}, we estimate optical \textit{flow-warped predictions} from each surrounding \textit{reference frame} to leverage temporal consistency estimation in pseudo-label selection (Fig.~\ref{f2}). Using these predictions, we selectively identify the most reliable pseudo-labels for object class and bounding boxes in the target \textit{key frame}. In particular, we introduce cross-IoU and cross-divergence based object pair matching across a group of predictions to search the most consistent bounding boxes and class labels, respectively. Moreover, to strike a balance between confirmation bias and uncertainty noise in pseudo-labels, we combine hard-class training with soft-label distillation based on their consistency. Our proposed SSVOD largely closes the performance gap between sparse and dense annotations by achieving around 98\% of supervised mAP with over 95\% sparsity in annotations (Fig.~\ref{f1}). Moreover, SSVOD provides average 7.5\% higher mAP than naive integration of video object detectors and SOTA semi-supervised image techniques (Table~\ref{t1}). Our contributions are summarized as follows.

\begin{itemize}
    \item We introduce a novel semi-supervised video object detection framework to tackle practical challenges of video object detection with sparse annotations.

    \item We propose a motion-aware robust pseudo class label and bounding box filtering approach by exploiting the temporal feature space of group of frames. 
    
    \item We present an extensive experimental study that shows significant performance improvements on ImageNet-VID, Epic-KITCHENS, and YouTube-VIS datasets. 
    
\end{itemize}

\section{Related Work}
\subsection{Video Object Detection}
Several supervised video object detectors have been explored over the years~\cite{query, yolov, wu2019selsa, fgfa}. Initial work has focused on post-processing of sequential video predictions with image-based object detectors~\cite{tcnn, tubelet}. {FGFA}~\cite{fgfa} first introduced flow-guided reference feature aggregation from nearby frames by utilizing motion maps. 
{SELSA}~\cite{wu2019selsa} replaced the motion-based aggregation by full sequence-level semantics with object cross-attention weights.
Later, {TROI}~\cite{troi} improved upon {SELSA} by integrating temporal alignment of region-of-interest (ROI) features using the surrounding frames.
{TF-blender} \cite{tfblend} proposed dense blending of the reference frame aggregation by considering each pair of surrounding frames.
Relation distillation networks~\cite{relation} distilled object relations from nearby reference frames to augment object features. Recently, an end-to-end transformer-based approach was introduced with sequential spatial and temporal processing~\cite{transformer}. However, the performance of all these approaches has been limited by the availability of the annotated frames from different temporal horizons of videos.

\subsection{Semi-Supervised Object Detection}
Recently, several semi-supervised image object detection techniques have been proposed. First, {STAC}~\cite{sohn2020detection} introduced a teacher-student framework for semi-supervised object detection that estimates the pseudo-labels with a pre-trained teacher. The {unbiased teacher}~\cite{liu2021unbiased} approach introduced an end-to-end approach to update the teacher as a moving average of the weights from the student. However, it only considered the pseudo class labels. {Soft-Teacher}~\cite{soft} introduced box-jittering and  modified confidence thresholding of the pseudo-labels. {Instant-teaching}~\cite{instant} proposed a co-rectify scheme by maintaining two models for generating the pseudo-labels. 
A single-stage detector based scheme was introduced in~\cite{unbiasedv2}. However, these static image-based detection approaches cannot exploit the temporal dynamics~\cite{fgfa, wu2019selsa} for overcoming video-specific challenges, such as deblurring, pose variations, and camera defocus.

Some of the prior work has focused on semi-supervised approaches for video. Misra \textit{et al.}~\cite{sv1} proposed a dynamic approach to gradually learn and accumulate the unknown objects from videos. Hu \textit{et al.}~\cite{sv2} introduced a pseudo-label propagation scheme on the detected objects in the unlabeled frames. 
However, these non end-to-end video object detection schemes are primarily built upon image-based detectors which cannot utilize the motion guided feature enhancement techniques developed in video object detectors.

\section{Methodology}

\subsection{Overview of Supervised Video Object Detection}

Supervised video object detection primarily focuses on aggregating surrounding context from nearby \textit{reference frames} to enhance detection of a target \textit{key frame}. 
In particular, supervised training requires annotations on \textit{key frames}, whereas \textit{reference frames} are the nearby frames primarily used for feature enhancement.  
With access to densely annotated \textit{key frames} over the video, supervised methods can learn temporal dynamics across different time-steps (Fig.~\ref{f3}\textcolor{blue}{a}). In contrast, having access to sparsely annotated frames in a video limits the temporal learning of supervised methods (Fig.~\ref{f3}\textcolor{blue}{b}).

Consider a dataset $\mathcal{D}$ consisting $M$ videos where $\mathcal{D} = \{V^1, V^2, \dots, V^M \}$. Each training video contains $N$ frames with $n_k$ annotated key frames and $n_r$ reference frames, such that $n_k + n_r = N$. The supervised training objective for a video object detection network $Z_\theta$ parameterized by $\theta$ can be expressed as
\begin{equation}
    \underset{\theta}{argmin} \sum_{m=1}^M \sum_{t=1}^{n_k} \mathcal{L}( Z_\theta (R_{m}^{t-i}, K_{m}^t, R_{m}^{t+i}); y_{m}^t),
\end{equation}
where $\mathcal{L}$ is a pre-defined loss function, $K_m^t, y_{m}^t $ denote the $t^{th}$ \textit{key frame} and corresponding annotation in the $m^{th}$ video, respectively, and $R_m^{t-i}$, $R_m^{t+i}$ represent the \textit{reference frame} at time $t-i$ and $t+i$, respectively. We note that here, and in the rest of the paper, boundary cases are taken care of by ensuring $(t-i)$ is always positive.

The supervised training objective is limited by the availability of annotated \textit{key frames}. Though annotations for \textit{reference frames} are not required, it is necessary to use the nearby \textit{reference frames} for proper feature enhancement. With a limited number of labeled \textit{key frames} in a particular video, the majority of the frames from other time-steps will remain unused in the supervised setting.

\subsection{Problem Formulation for SSVOD}
Instead of only relying on labeled \textit{key frames}, the semi-supervised approach targets robust pseudo-label generation for unlabeled \textit{key frames} throughout the video (Fig.~\textcolor{blue}{\ref{f3}c}). Therefore, training can be continued across different time-steps of the video utilizing the labeled and pseudo-labeled \textit{key frames}. Given that each video contains $n_k^u$ unlabeled \textit{key frames} and $n_k^l$ labeled \textit{key frames} such that $n_k^u + n_k^l = n_k$, the semi-supervised objective can be defined as follows
\begin{equation}
\begin{split}
   \underset{\theta}{\text{argmin}}  \left( \sum_{m=1}^M \sum_{t=1}^{n_k^l} \mathcal{L}( Z_\theta (R_m^{t-i}, K_m^t, R_m^{t+i}); y_{m}^t) \right. + \\
    \left. \sum_{m=1}^M \sum_{t=1}^{n_k^{u}} \mathcal{L}( Z_\theta (R_m^{t-i}, K_m^t, R_m^{t+i}); p_m^t) \right),
\end{split}
\end{equation}
where $p^t_m$ denotes the pseudo-label generated for the $t^{th}$ unlabeled \textit{key frame} from the $m^{th}$ video. Therefore, the proposed semi-supervised formulation ensures utilization of sufficient key-reference pairs across the video without being entirely limited by human annotations.

\subsection{SSVOD Overview}

Our proposed semi-supervised video object detection (SSVOD) framework facilitates training of SOTA video object detectors with sparse annotations (Fig.~\ref{f2}). To utilize the temporal information with video detectors, SSVOD operates on a group of frames to aggregate features of nearby \textit{reference frames} for \textit{key frame} prediction. For having sparsely annotated \textit{key frames}, SSVOD inherits the teacher-student framework that is extensively used in semi-supervised learning~\cite{soft, liu2021unbiased} for generating robust pseudo-labels on unlabeled \textit{key frames}. The teacher network is derived by the exponential moving average (EMA) weights of the student network which operates on weakly-augmented unlabeled sets to estimate pseudo-labels on unlabeled \textit{key frames}. The student network is trained on strongly augmented unlabeled sets using estimated pseudo-labels along with labeled sets. The performance of semi-supervised learning mostly depends on the correct estimation and filtering of pseudo-labels. SSVOD searches for the most consistent and robust pseudo-labels on unlabeled \textit{key frames} by exploiting the temporal feature space of nearby frames. 

\begin{figure}[t]
  \centering
   \includegraphics[width=0.9\linewidth]{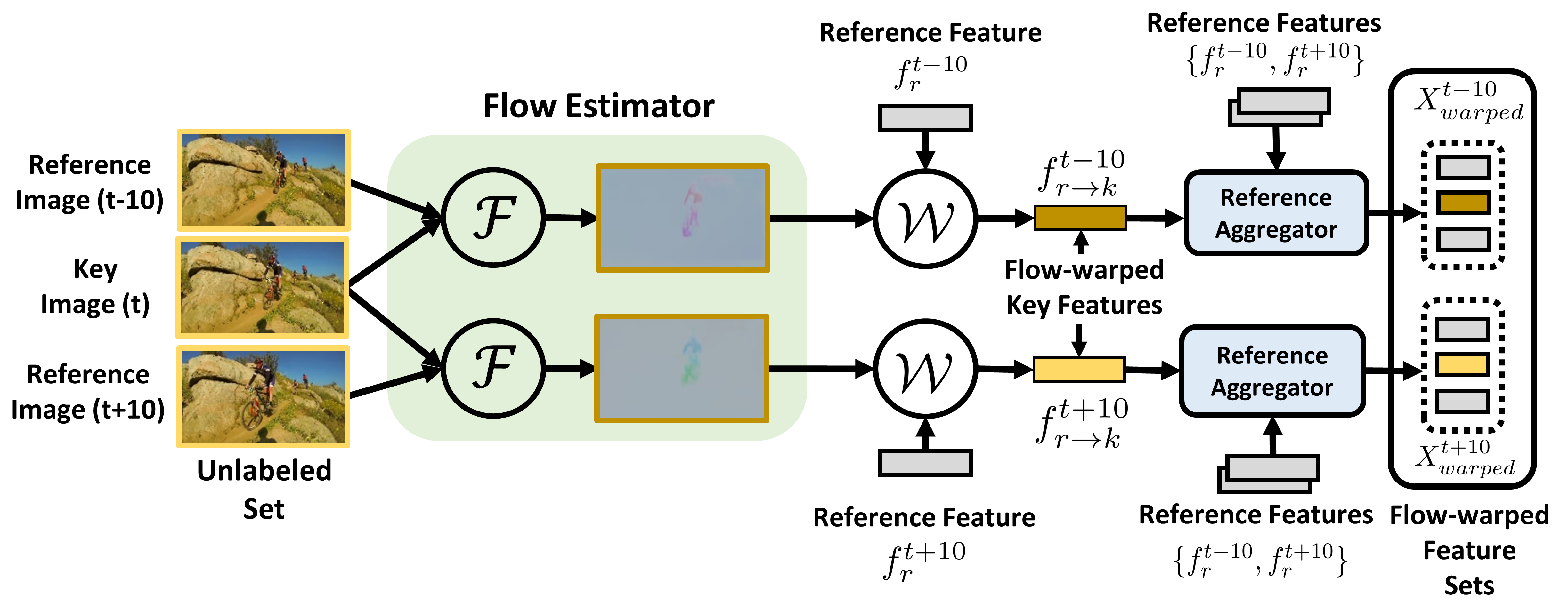}
   \caption{Overview of flow-warped feature set generation scheme. The flow estimator extracts the motion map between each key-reference pair. A feature warping is carried out on each \textit{reference feature}. The key feature is replaced with the flow-warped feature to generate \textit{flow-warped feature sets}.}
   \label{f4}
\end{figure}

Generally, in two-stage SOTA video detectors, the video feature extractor separately generates features from each \textit{key} and \textit{reference frame}. Later, the video region proposal network (RPN) and region-of-interest (ROI) head aggregates the reference features into the key feature to generate \textit{raw predictions} on \textit{key frames}. To leverage temporal consistency estimation on these \textit{raw predictions}, we introduce estimation of \textit{flow-warped predictions} from each \textit{reference frame} utilizing motion cues with a flow estimator. In this regard, we propose generation of \textit{flow-warped feature sets} from each key-reference pair 
based on the motion flow (Fig.~\ref{f4}, Sec~\ref{3.4}). These \textit{flow-warped feature sets} are processed simultaneously with the video RPN and ROI head to estimate \textit{flow-warped predictions} on the target \textit{key frame}. Therefore, instead of generating single \textit{key frame} predictions, SSVOD exploits the inherent temporal feature accumulation over a group of frames in the SOTA two-stage video detectors to estimate sets of predictions based on motion flow.

As a next step, we estimate temporal consistency across generated \textit{raw} and \textit{flow-warped predictions} to selectively filter robust pseudo bounding box and pseudo class labels on unlabeled \textit{key frames} (Sec~\ref{3.5}). For pseudo bounding box, we estimate cross-IoU (intersection over union) across each object pair in \textit{raw} and \textit{flow-warped predictions} to filter objects maintaining high regional overlapping. For pseudo class labels, we further estimate the cross-divergence in class predictions across object pairs with high regional overlapping to filter class labels with high consistency over different predictions. Soft-labels facilitate learning on inter-class relationships with more label-noise for higher uncertainty~\cite{humble, fixmatch}, whereas wrong estimation of hard-labels leads to confirmation bias~\cite{confirmation}.
To strike a balance between these two, we combine hard-label training with soft-label distillation based on measured class-label consistency.

\subsection{Flow-warped Feature Set Generation}
\label{3.4}
Inspired by FGFA~\cite{fgfa} which has introduced flow-guided feature aggregation in video object detection, we propose flow-warped feature set generation to estimate temporal prediction consistency. The flow-warped feature sets contain all the reference image features and the flow-warped key image features generated from motion warping of each of the reference image features. The details are presented in Figure~\ref{f4}. The $t^{th}$ raw feature set $X_{raw}^t$ having the key and reference image features of a particular video within range $[t-i, t+i]$ is denoted by
\begin{align}
X_{raw}^t = \{ f_r^{t-i}, \dots, f_k^t, \dots, f_r^{t+i} \},
\end{align}
where $f_r^{t-i}, f_k^t, f_r^{t+i}$ denote features generated from $R^{t-i}$, $K^{t}$, and $R^{t+i}$ image frames, respectively.

A flow network $\mathcal{F}(.)$ \cite{flownet} is used to generate the flow-map of each key-reference pair, and the feature warping function $\mathcal{W}(.)$ \cite{fgfa} is carried out on each reference image features using the estimated flow map to generate flow-warped key feature $f_{r \rightarrow k}$. Thus, the estimated flow-warped feature set ${X_{flow-warped}}$ from given \textit{key-reference features} is given by
\begin{align}
    & f_{r \rightarrow k}^{t+j} = \mathcal{W}(f_r^{t+j}, \mathcal{F}(K^t, R^{t+j})), \\
    & X_{flow-warped}^{t+j} = \{ f_r^{t-i}, \dots, f_{r \rightarrow k}^{t+j}, \dots, f_r^{t+i} \}, \\
    \nonumber &\forall j \ \in \{-i, \dots, i \}, j \ne 0.
\end{align}

In general, only two \textit{reference frames} are used with one \textit{key frame} in each feature set for training. Hence, the estimation of these sets is computationally efficient.

\subsection{Robust Pseudo-Label Selection}
\label{3.5}
On the target \textit{key frame}, we obtain {raw} $P_{raw}$ and flow-warped predictions $P_{flow-warped}$ after processing raw $X_{raw}$ and flow-warped feature sets $X_{flow-warped}$, respectively. To estimate the temporal consistency across these predictions for filtering robust pseudo bounding boxes, hard and soft class labels, we introduce three stages of selection following initial confidence thresholding.

\textbf{Cross-IoU based Pseudo Bounding Box Selection}.
As a first step, we estimate object pair matching between the raw and each flow-warped \textit{key frame} prediction based on maximum overlap. For the $k^{th}$ object in the $t^{th}$ frame $o^t_k$, the matched object in the $(t+j)^{th}$ flow-warped frame $o_k^{t+j}$ is estimated as
\begin{align}
    o_k^{t+j} = \underset{w \in \{1, \dots, n^{t+j} \}}{\text{argmax}}\ \text{IoU}(o_{k}^t, o_{w}^{t + j}), \\
    \nonumber \forall k \in \{1, \dots, n^t \}, j \in \{-i, \dots, i \}, j \ne 0,
\end{align}
where $n^{t}$, $n^{t+j}$ denotes number of objects at the $t^{th}$ \textit{key frame} and the $(t+j)^{th}$ \textit{reference frames}, respectively.
In the following, we estimate the mean cross-IoU (xIoU) for each object of the $t^{th}$ {raw prediction} $P_{raw}^t$ which represents the consistency in bounding boxes where 
\begin{align}
\resizebox{0.88\hsize}{!}{
    $\text{xIoU}(k) = \frac{1}{2i} \sum_{j=-i}^{i} \text{IoU}(o_k^t, o_k^{t+j} )\   \forall k \in \{1, \dots, n^t \}$.
    }
\end{align}
Finally, we filter out the objects with high xIoU scores to obtain the pseudo bounding boxes $P_{bbox}^t$, which is given by
\begin{equation}
    P_{bbox}^t = \mathbb{I}(\text{xIoU} > \zeta_{IoU}) P_{raw, bbox}^t,
\end{equation}
where $\mathbb{I}(.)$ is the indicator function, and $\zeta_{IoU}$ denotes the threshold IoU.

\begin{table*}[]
\centering
\caption{Performance comparison on ImageNet-VID validation set under \textbf{single labeled key frame per video} setting with different sampling rates and under \textbf{multiple labeled key frames per video} setting from $25\%$ training videos. * denotes our improved implementation for the VOD task. All the results are average of three independent runs. We reproduce state-of-the-art supervised video, semi-supervised image, and supervised image baseline schemes for benchmarking in the similar setting. No ImageNet-DET pre-training is considered.}
\label{t1}
\scalebox{0.8}{
\begin{tabular}{lccccccc}
\toprule
\multirow{2}{*}{Method} & \multicolumn{4}{c}{Single Labeled Key Frame (Percentage)} & \multicolumn{3}{c}{Multiple Labeled Key Frames (Number)}  \\ \cmidrule(r){2-5} \cmidrule(l){6-8}
                        & 25\%     & 50\%     & 75\%    & 100\%     & 5       & 10      & 15    \\
\hline
(a) Supervised Image Baseline~\cite{faster} & 30.1 $\pm$ 0.25 & 37.6 $\pm$ 0.22 & 42.5 $\pm$ 0.18 & 47.8 $\pm$ 0.29 & 35.2 $\pm$ 0.26 & 39.9 $\pm$ 0.26 & 43.1 $\pm$ 0.18 \\
(b) STAC~\cite{stac}                      & 37.2 $\pm$ 0.31 & 44.2 $\pm$ 0.18 & 46.6 $\pm$ 0.24 & 50.9 $\pm$ 0.19 & 39.5 $\pm$ 0.29 & 41.3 $\pm$ 0.19 & -\\
(c) Soft-Teacher~\cite{soft}              & 40.2 $\pm$ 0.26 & 46.7 $\pm$ 0.16 & 49.7 $\pm$ 0.18 & 54.7 $\pm$ 0.21 & 40.8 $\pm$ 0.32 & 42.2 $\pm$ 0.27 & -\\
(d) Unbiased Teacher~\cite{liu2021unbiased}          & 39.1 $\pm$ 0.16 & 44.6 $\pm$ 0.26 & 48.3 $\pm$ 0.21 & 53.4 $\pm$ 0.28 &40.1 $\pm$ 0.23 & 41.8 $\pm$ 0.18 & - \\ \hline
(e) Supervised Video Baseline~\cite{wu2019selsa} & 39.6 $\pm$ 0.23 & 43.1 $\pm$ 0.17 & 48.9 $\pm$ 0.19 & 55.0 $\pm$ 0.28 & 44.3 $\pm$ 0.32 & 47.4 $\pm$ 0.24 & 50.8 $\pm$ 0.28\\
(f) Baseline~\cite{wu2019selsa} + STAC~\cite{stac}              & 43.4 $\pm$ 0.26 & 47.9 $\pm$ 0.21 & 53.5 $\pm$ 0.18 & 57.4 $\pm$ 0.12 & 45.6 $\pm$ 0.21 & 48.3 $\pm$ 0.25 & - \\
(g) Baseline~\cite{wu2019selsa} + Soft-Teacher~\cite{soft}              & 45.2 $\pm$ 0.21 & 50.2 $\pm$ 0.22 & 55.8 $\pm$ 0.20 & 59.3 $\pm$ 0.19 & 46.4 $\pm$ 0.23 & 48.8 $\pm$ 0.19 & - \\
(h) Baseline~\cite{wu2019selsa} + Unbiased Teacher~\cite{liu2021unbiased}          & 44.7 $\pm$ 0.28 & 51.0 $\pm$ 0.25 & 55.1 $\pm$ 0.23 & 58.8 $\pm$ 0.17 & 46.2 $\pm$ 0.24 & 48.7 $\pm$ 0.18 & - \\
(i) PseudoProp~\cite{sv2} + Soft-Teacher~\cite{soft}          & 41.1 $\pm$ 0.26 & 47.6 $\pm$ 0.26 & 51.6 $\pm$ 0.21 & 56.5 $\pm$ 0.17 & 45.3 $\pm$ 0.27 & 48.0 $\pm$ 0.16 & - \\ 
(j) Misra \textit{et al.}*~\cite{sv1}              & 42.9 $\pm$ 0.23 & 48.5 $\pm$ 0.25 & 53.1 $\pm$ 0.19 & 57.7 $\pm$ 0.21  & 47.1 $\pm$ 0.22 & 49.2 $\pm$ 0.21 & - \\
(k) Yan \textit{et al.}~\cite{yan2019semi}*            & 40.5 $\pm$ 0.23 & 44.2 $\pm$ 0.18 & 50.3 $\pm$ 0.21 & 52.5 $\pm$ 0.26  & 46.2 $\pm$ 0.23 & 47.0 $\pm$ 0.27 & - \\
(l) Ours (SSVOD)                      & \textbf{48.6} $\pm$ 0.23 & \textbf{53.4} $\pm$ 0.18 & \textbf{60.5} $\pm$ 0.24 & \textbf{63.8} $\pm$ 0.19 & \textbf{49.7} $\pm$ 0.24 & \textbf{50.4} $\pm$ 0.22 & - \\
\bottomrule
\end{tabular}}
\end{table*}

\textbf{Cross-Divergence based Pseudo Hard Class Selection}.
In some cases, the bounding box estimation can be accurate, even though the object class is wrong. To address this issue, we estimate the mean cross KL-divergence (xDiv) for each object $o_k^t$ in the raw prediction $P_{raw}^t$ with its corresponding flow-warped object $o_k^{t+j}$, such that
\begin{align}
\resizebox{0.88\hsize}{!}{
    $\text{xDiv}(k) = \frac{1}{2i} \sum_{j=-i}^{i} D_{KL} (o_k^t, o_k^{t+j} ),\   \forall k \in \{1, \dots, n^t \}$.}
\end{align}
The bounding boxes with lower KL-divergence are filtered for hard class labeling with threshold $\eta_{div}$, such that
\begin{equation}
    P_{cls}^t = \mathbb{I}(\text{xDiV} < \eta_{div}) P_{raw, cls}^t.
\end{equation}

\textbf{Confidence-Aware Pseudo Soft Class Selection}.
To enhance the exploration of the generated pseudo-labels, class predictions from remaining bounding boxes in $P_{raw, cls}^t$  are accumulated after confidence thresholding with threshold $\gamma_c$ to generate the  soft-class distribution $P_{soft}^t$ , \textit{i.e.},
\begin{align}
\resizebox{.88\hsize}{!}{
    $P_{soft}^t = \mathbb{I}(c>\gamma_c; \text{xDiV} > \eta_{div}; \text{xIoU} < \zeta_{IoU}) P_{raw, cls}^t$.}
\end{align}
A soft-distillation is carried out between the class prediction of student network $P_{std, cls}^t$ and the filtered soft-class distribution from teacher $P_{soft}^t$ at time $t$,  such that
\begin{equation}
    \mathcal{L}_{soft}^t = D_{KL} (P_{std, cls}^t, P_{soft}^t).
\end{equation}
Finally, the objective loss to train the student is defined as
\begin{align}
    \mathcal{L} = \mathcal{L}_{cls}^{sup} + \mathcal{L}_{bbox}^{sup} + \mathcal{L}_{cls}^{unsup} + \mathcal{L}_{bbox}^{unsup} + \mathcal{L}_{soft}^{unsup}.
\end{align}

\section{Experiments}
\subsection{Dataset and Evaluation Setup}
Following prior work~\cite{fgfa, wu2019selsa, troi}, we primarily use the large-scale ImageNet-VID dataset~\cite{ILSVRC15} for experiments. The training set contains $3,862$ videos collected at a frame rate of $25$ frames per second (fps) to $30$ fps, and the validation set contains $555$ videos. A total of $30$ object categories are included, which is a subset of ImageNet-DET dataset containing $200$ categories of static images~\cite{ILSVRC15}. Existing supervised schemes used the ImageNet-DET dataset for pre-training which is a large-scale static-image dataset with similar objects. Since such pre-training on ImageNet-DET disrupts the sparse annotation constraints on the ImageNet-VID performance, we primarily focus on the ImageNet-VID dataset with sparse annotations. However, some experiments are done to analyze the effect of ImageNet-DET inclusion. For training, we considered the same $15$ uniformly sampled labeled \textit{key frames} per video as in prior work~\cite{fgfa, wu2019selsa, tfblend} with the following two settings.

\noindent \textbf{Single Labeled Key Frame per Video.}
Only one out of these $15$ key frames is used as a labeled \textit{key frame} per video while the remaining $14$ are used as unlabeled \textit{key frames} during training. We sample $25\%$, $50\%$, $75\%$, and $100\%$ videos with a single labeled \textit{key frame}. The nearby surrounding frames of these \textit{key frames} are considered reference frames where annotations are not required.

\noindent \textbf{Multiple Labeled Key Frames per Video.}
We sample $5$, $10$, and $15$ labeled \textit{key frames} from a total of $15$ per video to represent different degrees of annotation sparsity during training. 
Moreover, we sample $1,000$ videos $(\approx 25\%)$ from the whole training dataset. Following standard practice~\cite{wu2019selsa, fgfa}, nearby surrounding frames of labeled/unlabeled \textit{key frames} are used as \textit{reference frames}.

\noindent \textbf{Additional Datasets.}
Since ImageNet-VID lacks diversity of objects, we also study the performance on challenging Epic-KITCHENS~\cite{damen2020epic} dataset that contains $290$ classes from $272$ videos taken from $28$ kitchen environments. For additional comparisons on video object detection, we use YouTube-VIS~\cite{yang2019video} dataset having $40$ object categories with $2238$, $302$, and $343$ training, validation, and test videos.

Following prior works~\cite{fgfa, wu2019selsa}, we follow the same convention to report the results with the standard mAP evaluation metric. The categorization of different size objects and different motion objects follows the design in~\cite{fgfa}. All experiments are conducted with three different sets generated with independent sampling.
  
\subsection{Implementation Details}
We use the implementation and hyper-parameters based on MMTracking~\cite{mmtracking}. For each \textit{key frame}, we use two \textit{reference frames} in training and $30$ \textit{reference frames} in evaluation following prior work~\cite{fgfa, wu2019selsa, troi}. For the video object detector, we mostly focus on {SELSA} network~\cite{wu2019selsa} unless otherwise specified.
We use the COCO-pretrained Faster-RCNN network~\cite{faster} as the base object detection network with FPN~\cite{fpn} and ResNet-50~\cite{resnet} backbone. We conduct experiments on $16$ NVIDIA RTX A5000 GPUs. More details can be found in supplementary materials.

\subsection{Main Results}
In this section, we present the key results obtained with the proposed SSVOD framework. We also compare SSVOD with the image and video baselines. We report the mAP with IoU set to $50\%$, denoted as mAP@IoU=0.5, on the ImageNet-VID validation set. We note that each method is associated with a letter for clarity in reporting the results (\textit{e.g.}, SSVOD (l), see Table ~\ref{t1}).

\textbf{Single Labeled Key Frame per Video Performance.}
In this setup, supervised video baseline provides an average of $+7$  mAP performance improvement over image baseline, as shown in Table~\ref{t1}. SSVOD (l) significantly outperforms the supervised baselines by achieving $+46.4\%$ and $+22.8\%$ higher mAP than supervised image (a), and video (e) baselines, respectively. 
We study the effect of SOTA semi-supervised image techniques~\cite{stac, soft, liu2021unbiased} for pseudo-label filtering on image and video baselines. We notice considerable improvements over the supervised baseline with these techniques, \textit{e.g.}, Soft-Teacher~\cite{soft} improves the image baseline by average $+7.7$ mAP (c), and video baseline by $+5.8$ mAP (g). However, since these image techniques cannot exploit temporal dynamics of a group of frames in videos, they provide sub-optimal improvement over video baseline. By overcoming such limitations with motion-guided pseudo-label filtering, SSVOD achieves average $+4$ mAP higher than the best performing naive integration of video baseline and Soft-Teacher (g). Though PseudoProp~\cite{sv2} leverages motion propagation in pseudo-label estimation, its non end-to-end image-based feature processing primarily focuses on the post-processing of generated outputs rather than exploration of temporal feature space of groups of frames. Hence, it provides sub-optimal performance even when integrated with SOTA image techniques (i).
The iterative supervised training and object tracking approach introduced in (i) for generating pseudo-labels are time-consuming (requiring $30$ iterations) and fail to produce reliable pseudo-labels on distant frames due to its reliance on tracking performance. 
The method in (k) generates pseudo-labels by relying on annotated frames at regular intervals, leading to incorrect pseudo-label estimation on distant frames under high sparsity of annotations.

\begin{table}[t]
\centering
\caption{Ablation with various video object detector. Our SSVOD scheme provides consistent improvements over supervised baseline across various detectors.}
\label{t3}
\scalebox{0.75}{
\begin{tabular}{lcccc}
\hline
detector               & method     & mAP                     & mAP@0.5                   & mAP@0.75    \\ \hline
\multirow{2}{*}{FGFA~\cite{fgfa}}  & Supervised & 23.4                     & 51.3                      & 28.8        \\
                       & Ours       & \textbf{31.9} (\color{blue}+8.5)   & \textbf{58.7} (\color{blue}+7.4)   & \textbf{36.4} (\color{blue}+7.6)  \\ \hline
\multirow{2}{*}{SELSA~\cite{wu2019selsa}} & Supervised & 32.1                      & 55.0        & 34.7                        \\
                       & Ours       & \textbf{39.2} (\color{blue}+9.2)   & \textbf{63.8} (\color{blue}+8.8)    & \textbf{43.1} (\color{blue}+8.4)  \\ \hline
\multirow{2}{*}{TROI~\cite{troi}}  & Supervised & 33.9                      & 56.2        & 37.1                        \\
                       & Ours       & \textbf{42.6} (\color{blue}+8.7)    & \textbf{64.8} (\color{blue}+8.6)    & \textbf{44.8} (\color{blue}+7.7)  \\ \hline
\end{tabular}}
\end{table}

\begin{table}[t]
\centering
\caption{Ablation study on the effect of different loss components on pseudo-labels.}
\label{t6}
\scalebox{0.8}{
\begin{tabular}{cccccc}
\hline
\multirow{2}{*}{\begin{tabular}[c]{@{}c@{}}Hard\\ class\end{tabular}} & \multirow{2}{*}{\begin{tabular}[c]{@{}c@{}}Bounding\\ box\end{tabular}} & \multirow{2}{*}{\begin{tabular}[c]{@{}c@{}}Soft\\ class\end{tabular}} & \multirow{2}{*}{mAP} & \multirow{2}{*}{mAP@0.5} & \multirow{2}{*}{mAP@0.75} \\
                                                                      &                                                                         &                                                                       &                      &                          &                           \\ \hline
\                                                                     &                                                                         &                                                                       & 32.1                 & 55.0                     & 34.7                      \\ 
\checkmark                                                                     &                                                                         &                                                                       & 35.6                 & 58.8                     & 37.9                      \\ 
\checkmark                                                                     & \checkmark                                                                       &                                                                       & 37.9                 & 61.1                     & 41.7                      \\ 
\checkmark                                                                     & \checkmark                                                                       & \checkmark                                                                     & \textbf{39.2}                 & \textbf{63.8}                     & \textbf{43.1}                      \\ \hline
\end{tabular}}
\end{table}

\begin{table}[]
\centering
\caption{Ablation study on the effect of different number of unlabeled \textit{key frames}.}
\label{t5}
\scalebox{0.8}{
\begin{tabular}{cccc}
\hline
\multirow{2}{*}{\begin{tabular}[c]{@{}c@{}}\# of unlabeled \\ key frames\end{tabular}} & \multirow{2}{*}{mAP} & \multirow{2}{*}{mAP@0.50} & \multirow{2}{*}{mAP@0.75} \\ 
                                                                                    &                               &                           &                           \\ \hline
1                                                                                   & 34.7                          & 54.5                      & 38.8                      \\
5                                                                                   & 37.1                          & 59.4                      & 41.6                      \\
10                                                                                  & 38.6                          & 62.1                      & 42.7                      \\
14                                                                                  & \textbf{39.2}                          & \textbf{63.8}                      & \textbf{43.1}                       \\ \hline                     
\end{tabular}}
\end{table}

\begin{table}[]
\centering
\caption{The effect of ImageNet-DET dataset integrated as the unlabeled set using single key frame from ImageNet-VID dataset in the labeled set.}
\label{t4}
\scalebox{0.85}{
\begin{tabular}{cccc}
\hline
Unlabeled Dataset                  & mAP       \\ \hline
ImageNet-VID (14 Key)               & 39.2       \\ 
ImageNet-VID (14 Key) + ImageNet-DET & \textbf{41.1} (\color{blue}+1.9)  \\ \hline
\end{tabular}}
\end{table}

\begin{table}[]
\centering
\caption{Additional study with SOTA VOD and flow estimators on SSVOD framework in the single-frame (100\% videos) setting.}
\label{tadd}
\scalebox{0.8}{\begin{tabular}{ccc}
\toprule
\textbf{VOD} & \textbf{Flow Estimator} & \textbf{ImageNet-VID} \\
\hline
SELSA~\cite{wu2019selsa}        & FlowNet~\cite{flownet}                 & 63.8 $\pm$ 0.19           \\
SELSA~\cite{wu2019selsa}        & FlowFormer~\cite{huang2022flowformer}              & 68.4 $\pm$ 0.18             \\
TransVOD~\cite{zhou2022transvod}     & FlowNet~\cite{flownet}                 & 67.2 $\pm$ 0.23           \\
TransVOD~\cite{zhou2022transvod}     & FlowFormer~\cite{huang2022flowformer}              & \textbf{70.5} $\pm$ 0.26          \\
\bottomrule
\end{tabular}}
\end{table}

\begin{table}[]
\centering
\caption{Additional comparisons on Epic-KITCHENS and YouTube-VIS Datasets on single-frame (100\% videos) and multi-frame (5 frames per video on 25\% videos) settings.}
\label{t9}
\scalebox{0.65}{
\begin{tabular}{cccccc}
\hline
\multirow{2}{*}{\textbf{Method/Dataset}} & \multicolumn{2}{c}{EPIC-KITCHENS~\cite{damen2020epic}} & \multicolumn{2}{c}{YouTube-VIS~\cite{yang2019video}} \\
\cmidrule(lr){2-3} \cmidrule(lr){4-5}

 & \textbf{Single}          & \textbf{Multi}         & \textbf{Single}         & \textbf{Multi}         \\
                                 \hline
Supervised~\cite{wu2019selsa}         & 30.4 $\pm$ 0.18              & 25.8 $\pm$ 0.26            & 45.3 $\pm$ 0.29             & 38.8 $\pm$ 0.15            \\
PseudoProp~\cite{sv2}   & 32.4 $\pm$ 0.29              & 29.2 $\pm$ 0.24            & 47.6 $\pm$ 0.17             & 41.8 $\pm$ 0.23            \\
Misra et al.~\cite{sv1}*   & 31.6 $\pm$ 0.23              & 28.1 $\pm$ 0.22            & 48.9 $\pm$ 0.19             & 42.2 $\pm$ 0.17            \\
Yan et al.~\cite{yan2019semi}*      & 29.2 $\pm$ 0.20              & 26.7 $\pm$ 0.24            & 43.7 $\pm$ 0.25             & 41.4 $\pm$ 0.23            \\
SSVOD (ours)            & $\mathbf{36.7 \pm 0.22}$              & $\mathbf{30.9 \pm 0.23}$            & $\mathbf{53.6 \pm 0.21}$             & $\mathbf{45.1 \pm 0.21}$           \\
\hline
\end{tabular}}
\vspace{-10pt}
\end{table}

\textbf{Multiple Labeled Key Frames per Video Performance.}
In this setting, we observe consistent performance improvement with the increase in labeled key frames per video for all baselines, as shown in Table~\ref{t1}. For the supervised video baseline (e), we notice $+4.7$, $+7.8$, $+11.2$ increase in mAP with the increase of labeled \textit{key frames} from $1$ to $5$, $10$, and $15$, respectively. This shows that supervised performance greatly depends upon the availability of labeled \textit{key frames} per video.
In general, the supervised video baselines (e) maintain superior performance over image baselines (a) and inclusion of semi-supervised image techniques considerably improve the supervised performance as before. However, our SSVOD (l) outperforms all the baselines by a considerable margin and largely closes the gap between sparse training and full-supervised training, achieved with all $15$ labeled \textit{key frames} per video.

\subsection{Ablation Studies}
In this section, we validate the effect of each design choice and parameter setting. All the ablation studies are conducted on the full training set with a single labeled key frame per video, unless specified otherwise. More ablations and visualizations are provided in the appendix.

\textbf{Effect of Different Video Object Detectors.}
Though we used SELSA~\cite{wu2019selsa} as a proof-of-concept for most experiments, we ablate the effect of different video object detectors on the SSVOD framework as shown in Table~\ref{t3}. Among supervised schemes, SELSA detector~\cite{wu2019selsa} provides superior performance to FGFA~\cite{fgfa}, while TROI~\cite{troi} exhibits the best performance. The SSVOD performance improvements are consistent with its supervised counterpart across different detectors. TROI detector achieves the best performance with $56.2$ mAP in supervised setting and its  performance reaches $64.8$ mAP for SSVOD, which is $+8.6$ points higher. These results demonstrate that SSVOD can scale-up the performance with improved supervised baselines.

\textbf{Effect of Different Loss Components on Pseudo-Labels.}
We study the effect of different loss components on pseudo-labels (see Table~\ref{t6}). Integrating hard-label cross-entropy loss improves performance by $+3$ points. Further applying pseudo bounding box regression loss, we get $+2.3$ mAP improvements. Finally, by incorporating soft-label distillation, we achieve the best performance of $63.8$ mAP that shows the effectiveness of the three loss components.

\begin{figure}[t]
  \centering
   \includegraphics[width=0.8\linewidth]{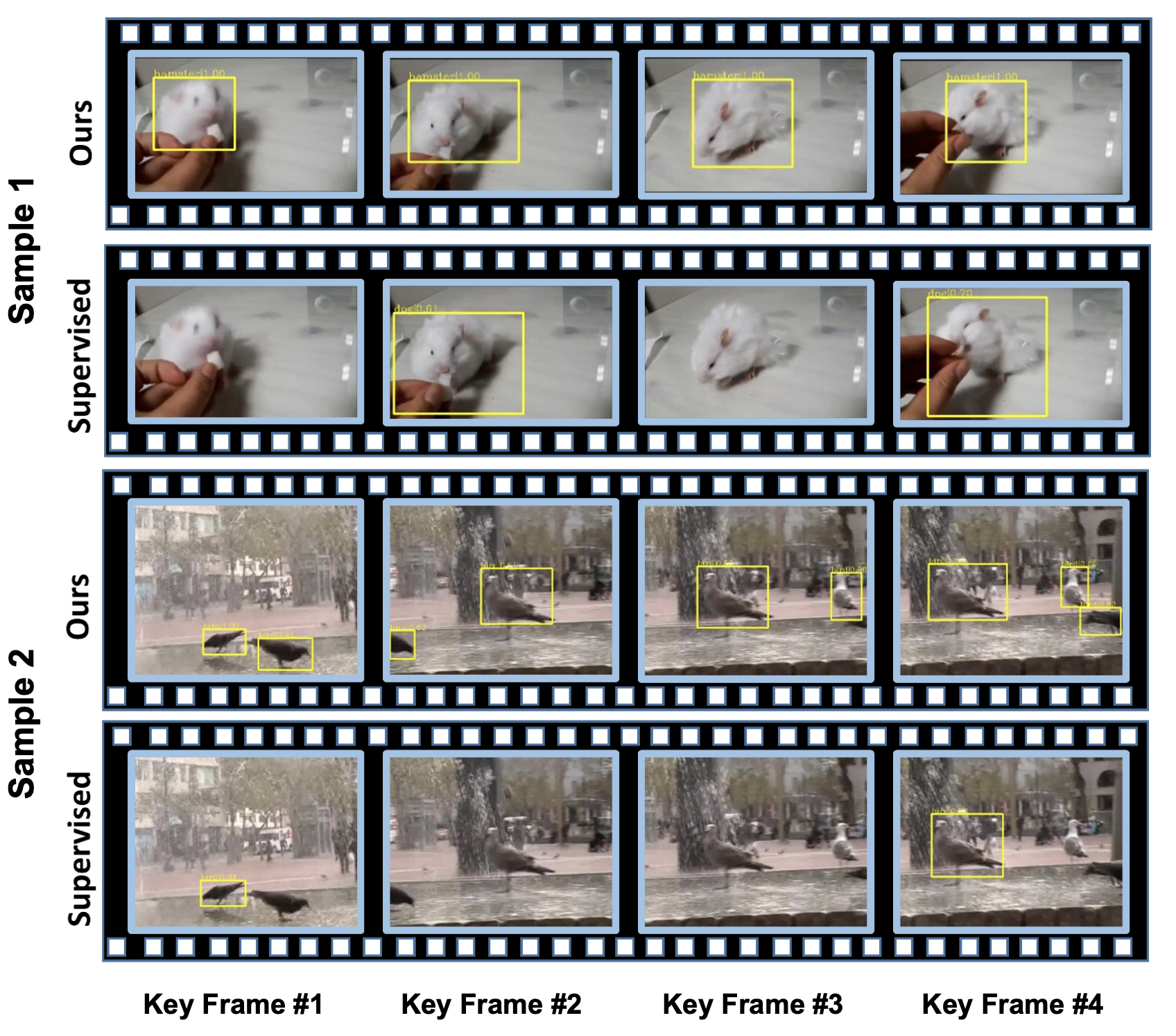}
   \caption{
   Qualitative visualization of supervised and SSVOD performance: SSVOD demonstrates better temporal consistency in predictions compared to its supervised counterpart over both challenging single and multi-object scenarios.}
   \label{f5}
\end{figure}

\textbf{Number of Unlabeled \textit{Key Frames}.}
We evaluate the performance for different numbers of unlabeled \textit{key frames}.
The performance consistently improves with the increasing number of unlabeled \textit{key frames}, as shown in Table~\ref{t5}. When a higher number of \textit{key frames} is used, SSVOD framework can explore different temporal regions of the video thereby producing better performance. We argue that SSVOD makes the annotations of multiple \textit{key frames} per video redundant. With one labeled and $14$ unlabeled \textit{key frames}, the SSVOD achieves $63.8$ mAP while the supervised baseline with $15$ labeled \textit{key frames} is at $64.7$ mAP.

\textbf{Correctness of Generated Pseudo-Labels.}
We study pseudo-label mAP on single labeled key frame per video setting with 25\% training videos: (See Table~\ref{t1} for naming) Method (f) is $57.1$, (g) is $59.8$, (h) is $58.5$, (i) is $55.2$, and ours (j) is $64.9$. The higher mAP achieved by SSVOD demonstrates better quality of pseudo-label generation.

\textbf{Effect of ImageNet-DET as Unlabeled Set.}
As described earlier, ImageNet-VID contains $30$ objects which are a subset of ImageNet-DET with $200$ object classes. Existing supervised schemes pre-train the video detector on the ImageNet-DET dataset that provides large performance gains~\cite{fgfa, troi,wu2019selsa}. Since ImageNet-DET is a static image dataset, video pre-training is conducted by considering the same image as both \textit{key} and \textit{reference frames}. However, such pre-training violates the constraints introduced by sparse annotations. Instead of such pre-training, we study the effects of ImageNet-DET integration as an unlabeled set. The results are given in Table~\ref{t4}. Despite having several unseen classes, we notice considerable performance improvement with the integration of ImageNet-DET. 

\textbf{Effect of SOTA VOD and flow estimators in SSVOD.}
We use FlowNet~\cite{flownet}, and baseline VODs~\cite{wu2019selsa} as proof-of-concept. We ablate SOTA FlowFormer~\cite{huang2022flowformer} and TransVOD~\cite{zhou2022transvod} in SSVOD (Table~\ref{tadd}). Integration of superior baseline models in SSVOD significantly improves performance, which is consistent with our prior observations.

\textbf{Performance on additional datasets.}
We present results on two additional benchmark datasets (EPIC-KITCHENS~\cite{damen2020epic} and Youtube-VIS~\cite{yang2019video}) for VOD tasks  (See Table~\ref{t9}). 
SSVOD consistently achieves superior performance on these benchmark datasets over existing methods.

\textbf{Qualitative Visualization of Performance:}
We study the qualitative performance of SSVOD and its corresponding supervised video baseline on challenging single and multi-object videos, as shown in Fig.~\ref{f5} (see appendix for more visualizations). SSVOD demonstrates visibly better temporal consistency than its supervised counterparts. In sample $1$, the supervised baseline wrongly classifies fast-moving \textit{hamster} as a \textit{dog} with much lower confidence, whereas SSVOD generates correct predictions consistently. In sample $2$, SSVOD consistently detects multiple objects in subsequent frames whereas the supervised baseline generates inconsistent predictions due to insufficient temporal exploration in training, that demonstrates its effectiveness.

\section{Conclusion}
In this paper, we introduce a novel semi-supervised learning framework (SSVOD) to overcome the limitations of existing supervised video object detection approaches for sparse annotations. Instead of the naive integration of existing semi-supervised image techniques on SOTA video detectors, SSVOD exploits temporal feature space of groups of frames to search robust pseudo-labels based on motion consistency. Moreover, SSVOD is found to be detector invariant which can scale-up performance with improved supervised baselines. Our proposed three-stage pseudo-label selection for bounding-box regression, hard-label classification, and soft-label distillation largely contributes to the final performance gain. Through effective utilization of unlabeled frames in videos, SSVOD achieves around 98\% of densely supervised performance by using over 95\% sparser annotations significantly outperforming other baselines.

{\small
\bibliographystyle{ieee_fullname}
\bibliography{ref}
}
\newpage

\appendix
\begin{center}
{\bf \Large  Appendix}
\end{center}
\bigskip 

\section{Overview}
In this supplementary material, we provide additional details and experimental results for our proposed SSVOD. In summary, the following items are presented.

\begin{itemize}
    \item Details of the mathematical notations used in the main paper.
    \item Details of the data augmentations used in model training.
    \item Additional ablation studies and performance analyses. 
    \item Analysis of the training loss curves for SSVOD.
    \item Additional qualitative comparisons on different class of objects.
\end{itemize}

\section{List of Notations}
All the notations used in the main paper are summarized in Table~\ref{t1}.

\section{Additional Implementation Details}
Additional details on the implementations are summarized in Table~\ref{t2}.

\section{Data Augmentations}
In SSVOD, we use three sets of augmentations to process the labeled set, unlabeled set for the teacher, and unlabeled set for the student, respectively. The details of the augmentations used in SSVOD is presented in Table~\ref{t3}. For strong augmentation, we consider additional geometric augmentations and the cutout augmentation~\cite{cutout}. In contrast, weak augmentation only contains random flipping to reduce data variations for the teacher network to enable more confident pseudo-label generation. We primarily incorporate the augmentation schemes that are heavily used in semi-supervised image object detection~\cite{soft, stac}. In SSVOD, the same augmentation parameters are maintained over each sequence of \textit{key} and \textit{reference} images in labeled and unlabeled sets. Since the \textit{reference frames} are mostly used to enhance the \textit{key} frame features, it is necessary to maintain the same augmentation choices on each set.

\section{Additional Ablations}
We perform ablation studies to compare the performance of the proposed SSVOD and baseline supervised approaches. We follow the single labeled \textit{key frame} per video setting for the supervised baseline and the proposed SSVOD approach. All cases are evaluated on the ImageNet-VID~\cite{ILSVRC15} validation set and we report the mAP@0.5 score unless otherwise mentioned.

\subsection{Per-Class Performance Analysis}
\begin{figure*}[]
        \centering
        \begin{subfigure}{.6\linewidth}
            \centering
            \caption{Supervised}
            \includegraphics[width=\textwidth]{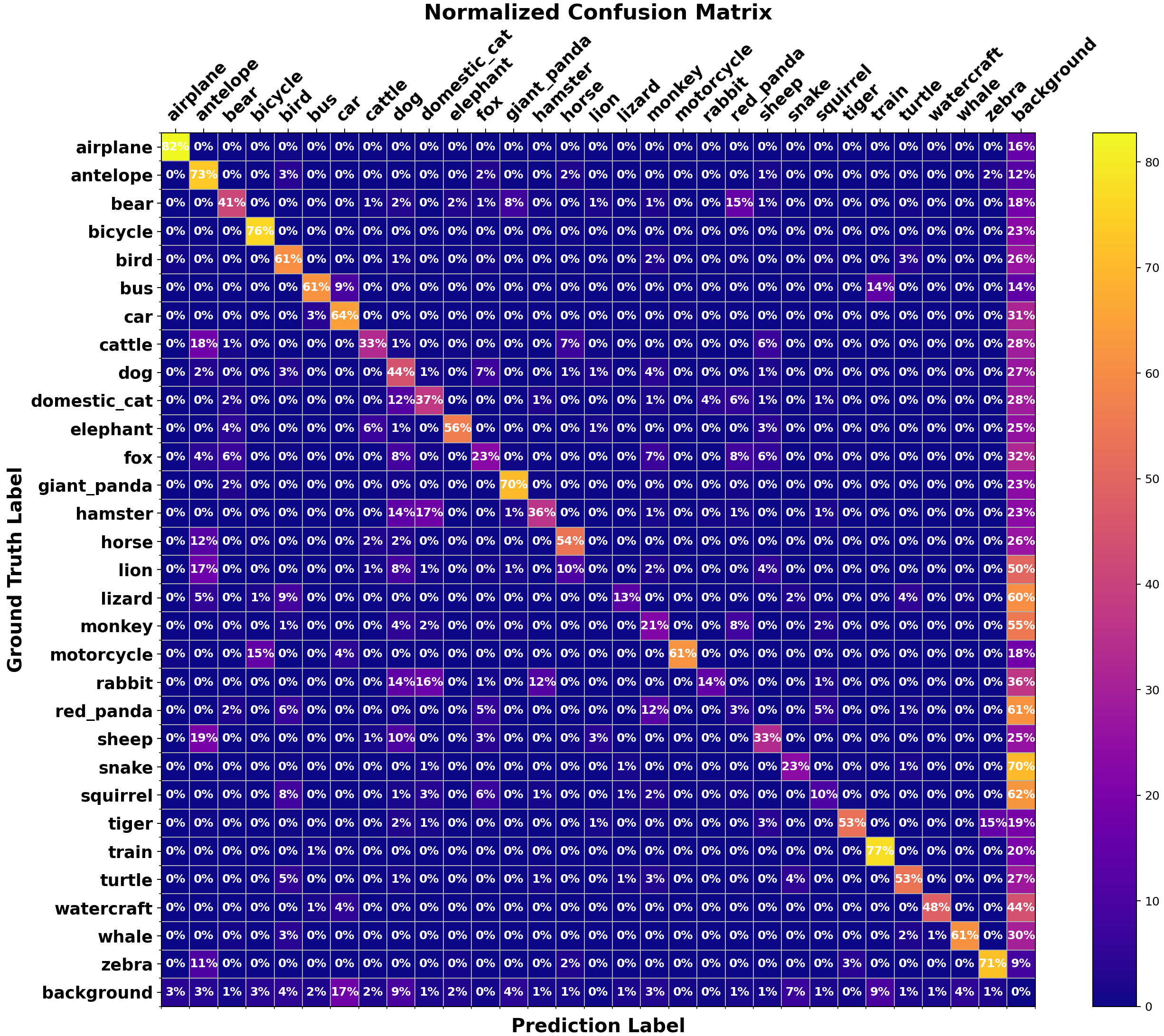}
            \begin{minipage}{1.4cm}
            \vfill
            \end{minipage}
        \end{subfigure}
        \vfill
        
        \begin{subfigure}{.6\linewidth}
            \centering
            \caption{SSVOD}
            \includegraphics[width=\textwidth]{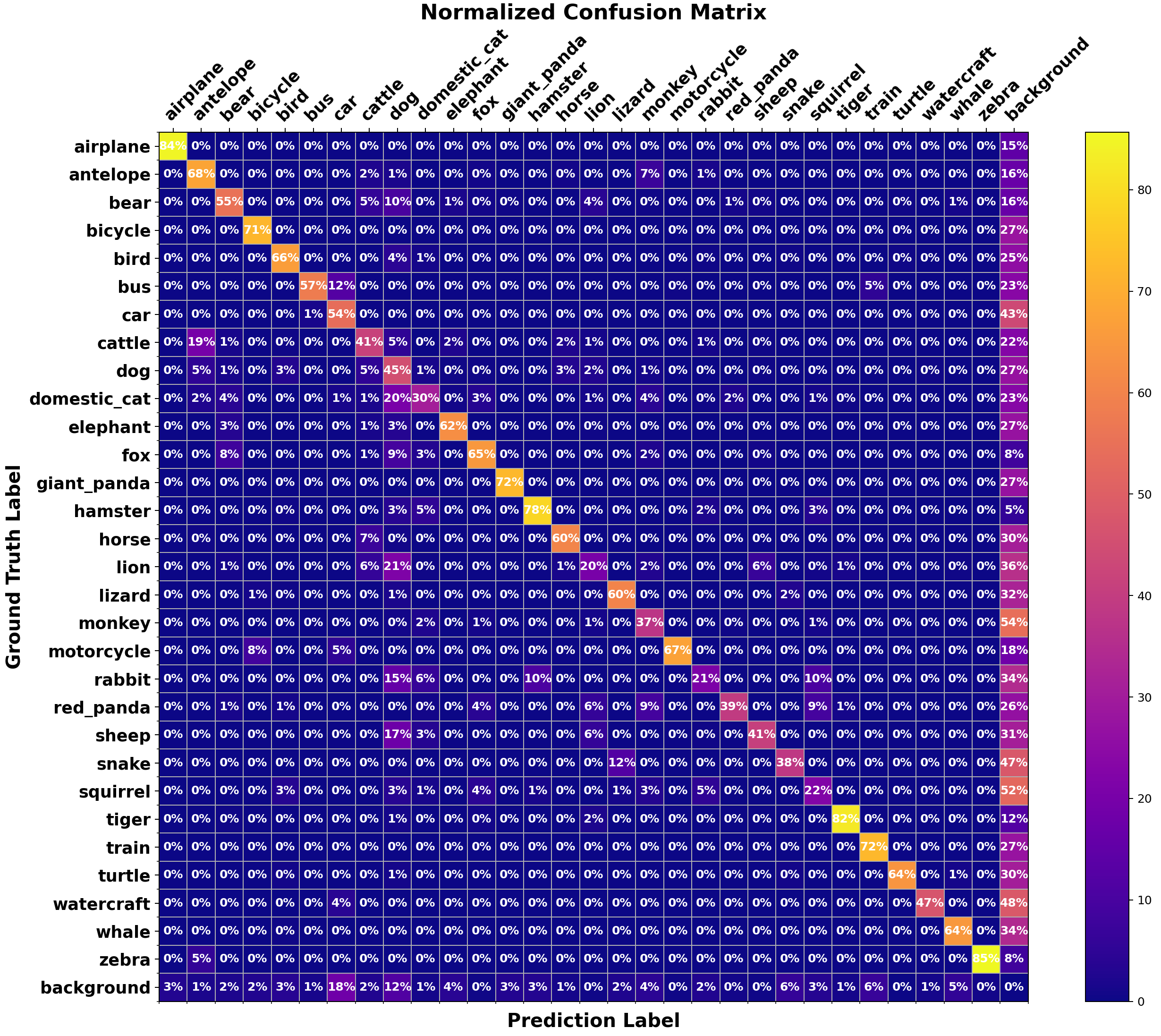}
            \begin{minipage}{.1cm}
            \vfill
            \end{minipage}
        \end{subfigure} 
        \caption{The normalized confusion matrix for the per-class classification performance with the (a) supervised, and (b) our proposed SSVOD approaches. SSVOD significantly improves the classification performance on the challenging classes in which the supervised approach performs poorly.}
        \label{confusion}
\end{figure*}

We study the per-class performance to determine the improvement gain on the evaluation set. In Figure~\ref{f5}, we present per-class mAP scores obtained from the baseline supervised approach and the proposed SSVOD scheme. SSVOD significantly improves the mAP score for most of the classes. We observe large mAP improvements on the challenging classes such as \textit{red panda} in which SSVOD achieves around $28$ times higher mAP. We further present normalized confusion matrix on the class predictions of supervised and SSVOD approaches, which is shown in Figure~\ref{confusion}.
Our SSVOD approach achieves consistent performances on most classes by considerably improving the performance for the challenging classes. For example, the minimum accuracy on \textit{lion} class with the supervised approach is nearly $0\%$, whereas with the SSVOD approach it rises to almost $20\%$. Thus, proper utilization of the unlabeled frames throughout the training video with the proposed SSVOD approach has great potential to improve the performance with scarce annotations.

\subsection{Loss Curve Analysis}
\begin{figure*}[]
        \centering
        \begin{subfigure}{.65\linewidth}
            \centering
            \caption{}
            \includegraphics[width=\textwidth]{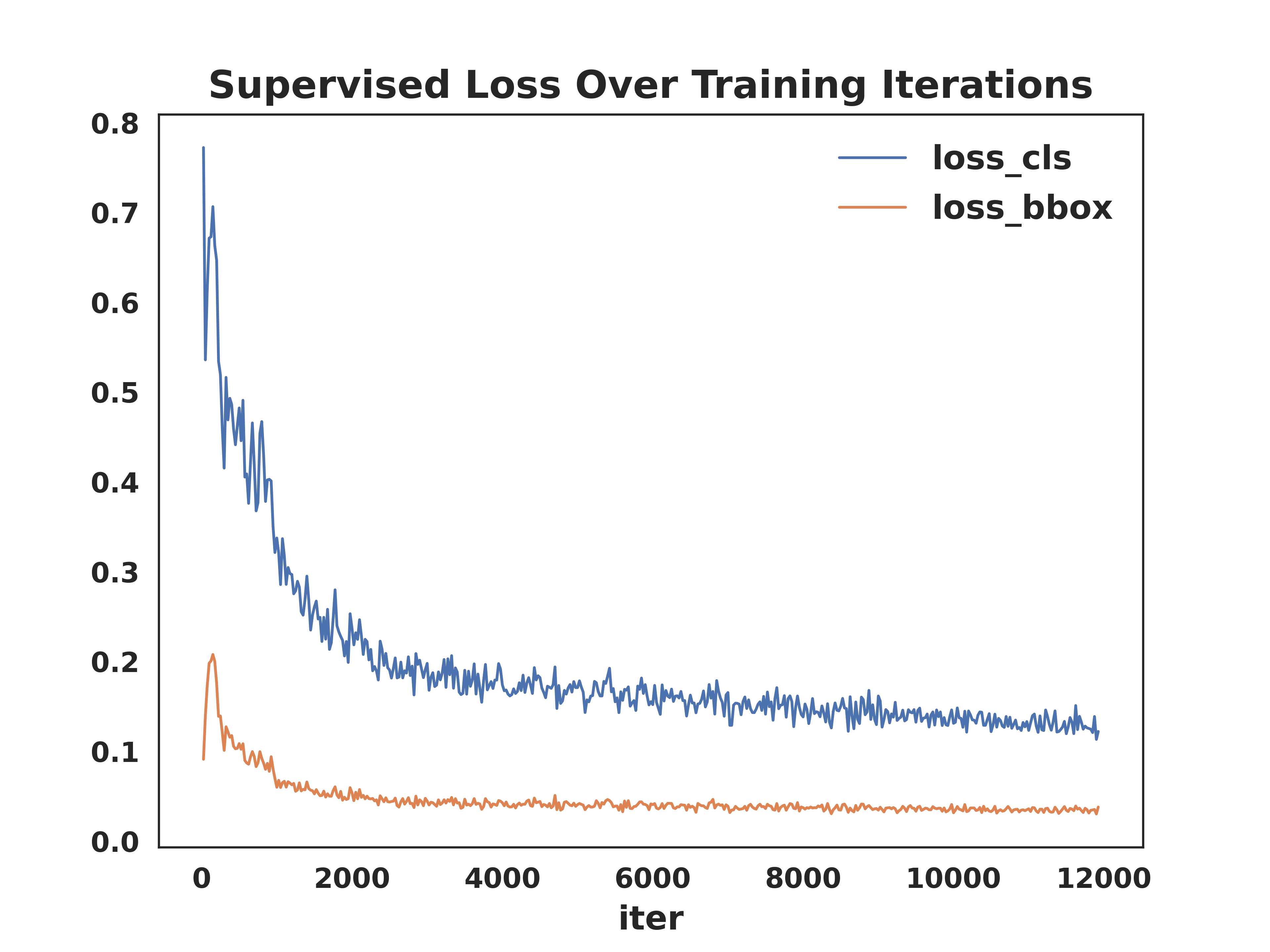}
            \begin{minipage}{4cm}
            \vfill
            \end{minipage}
        \end{subfigure}
        
        \vfill
        \begin{subfigure}{.65\linewidth}
            \centering
            \caption{}
            \includegraphics[width=\textwidth]{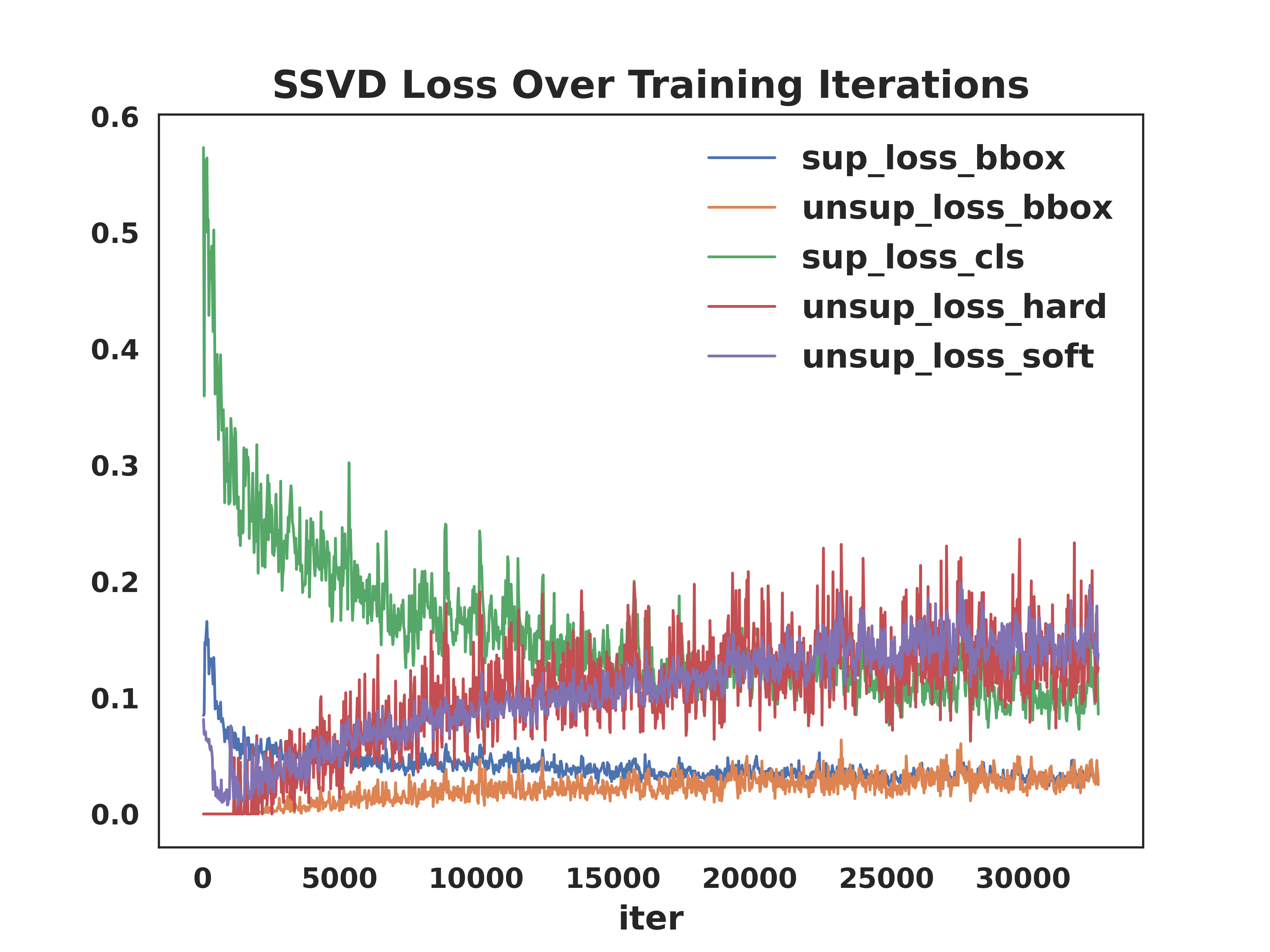}
        \end{subfigure} 
        \caption{Analysis of the loss curves with (a) supervised approach, and (b) our proposed SSVOD approach. Supervised losses gradually decreases and saturates. In SSVOD, unsupervised losses rises gradually with improved pseudo-label generation. Finally it converges with the supervised loss.}
        \label{loss}
\end{figure*}

We present loss curves over training iterations as shown in Figure~\ref{loss}. In the supervised training, both the classification and bounding box losses gradually decrease until saturation. In the SSVOD training, we notice a similar behavior on the supervised losses. For unsupervised losses in SSVOD, the hard classification loss and bounding box loss start from zero since the model can't generate high confident pseudo-labels which leads to filtration of all the labels. On the other hand, the unsupervised soft loss initiates the training on the unlabeled sets. Gradually, the model generates highly confident pseudo-labels and the unsupervised losses continue to rise. Finally, both the supervised and unsupervised losses converge together as training progresses.

\subsection{Performance Comparison on Objects of Different Sizes and Motions.}
We study the performance on objects with different sizes and motion categories following~\cite{fgfa}. The results are shown in Table~\ref{t8}. Recognizing smaller and faster objects is relatively more challenging. We observe that supervised accuracy is considerably low on small and fast objects. We achieve $+4.0$ and $+9.2$ mAP (@0.5:0.95) improvements on the middle and large objects, respectively. Accordingly, on the medium and slow objects, we notice consistent improvements of $+4.7$ and $+10.1$ points, respectively. However, the performance improvements are comparably smaller in the challenging small ($+2.1$ points) and fast ($+2.3$ points) objects.

\subsection{Effect of Different Choices of Pseudo-Label Thresholds on performance.}
We study the effect of different choices of thresholds in pseudo-label selection. We present the performance for combinations of threshold choices in Table~\ref{t7}. Best performance is achieved when the confidence threshold ($\gamma_c$) is  set to $0.8$ for soft class distillation, $0.9$ for IoU threshold ($\zeta_{IoU}$) in bounding box regression, and $0.005$ for KL-divergence threshold ($\eta_{div}$) in hard-label classification. Higher values for these thresholds reduce the number of pseudo-labels for the unlabeled set, whereas lower values compromise the pseudo-label quality, resulting in confirmation bias~\cite{instant}. We have carried out all other experiments with the best choice of these thresholds.

\subsection{Additional Qualitative Results}
\begin{figure*}[t]
  \centering
  \includegraphics[width=1\linewidth]{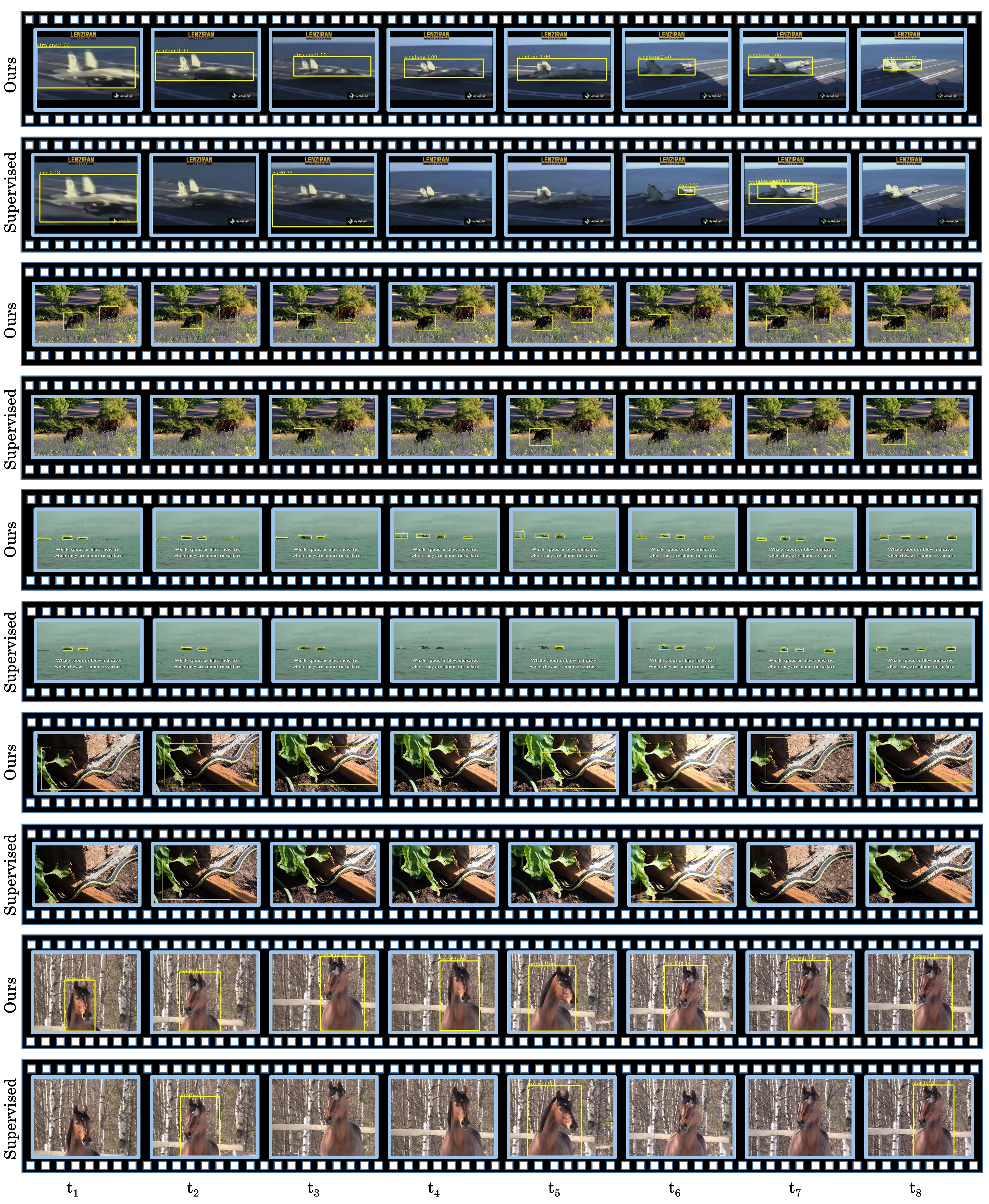} 
   \caption{Qualitative performance analysis between the supervised and our proposed SSVOD approach. SSVOD generates more consistent predictions over various time steps across the video.}
   \label{qual}
\end{figure*}

We provide additional visualizations of the qualitative results obtained from SSVOD and the supervised scheme in Figure~\ref{qual}. As discussed earlier, SSVOD can learn the temporal variations of each object by utilizing sets of unlabeled images from different time steps in a video. We observe that SSVOD generates more stable class and bounding box predictions throughout the video without any post-processing. In contrast, the supervised approach generates inconsistent results which can not properly adapt to the temporal variations present in challenging conditions.

\begin{table*}[]
\centering
\caption{Different notations used to describe the operations of SSVOD. We categorize the notations into three types.}
\label{t1}
\scalebox{1}{
\begin{tabular}{c|c|c}
\hline
Type              & Description                                 & Notation                                            \\ \hline
Scaler Parameters & number of videos                            & $M$                                                   \\
                  & number of frames per video                            & $N$                                                   \\
                  & number of \textit{key frames} per video                        & $n_k$                                                \\
                  & number of \textit{reference frames} per video                  & $n_r$                                                \\
                  & number of labeled \textit{key frames} per video                & $n_k^l$                             \\
                  & number of unlabeled \textit{key frames} per video              & $n_k^u$                             \\
                  & number of objects in the $t^{th}$ frame         & $n^t$                               \\
                  & cross-IoU threshold                         & $\zeta_{IoU}$                       \\
                  & confidence threshold                        & $\gamma_c$                            \\
                  & cross-divergence threshold                  & $\eta_{div}$                        \\ \hline
Functions/Models  & video object detection network              & $Z_\theta (\cdot)$  \\
                  & flow network                                & $\mathcal{F} (\cdot)$  \\
                  & feature warping                             & $\mathcal{W} (\cdot)$ \\
                  & cross-IoU estimator                         & $IoU(\cdot)$                           \\
                  & cross-KL divergence estimator               & $D_{KL}(\cdot)$                     \\ \hline
Vectors/Matrix    & \textit{key frame} at timestamp $t$ in the $m^{th}$ video               & $K_m^t$                             \\
                  & \textit{key frame} at timestamp $t-i$ in the $m^{th}$ video             & $R_m^{t-i}$                       \\
                  & annotations of the $t^{th}$ frame in the $m^{th}$ video  & $y_m^t$                             \\
                  & pseudo-label of the $t^{th}$ frame in the $m^{th}$ video & $p_m^t$                             \\
                  & \textit{key} feature                                 & $f_k$                                                \\
                  & \textit{reference} feature                           & $f_r$                                                \\
                  & flow-warped \textit{key} feature from the \textit{reference} frame                          & $f_{r \rightarrow k}$               \\
                  & Raw feature set at timestamp $t$                      & $X_{raw}^t$                       \\
                  & flow-warped feature set at timestamp ${t+j}$               & $X_{warped}^{t+j}$              \\
                  & predictions on raw feature set at timestamp $t$       & $P_{raw}^t$                       \\
                  & class predictions on raw feature set at timestamp $t$       & $P_{raw, cls}^t$                       \\
                  & bounding box predictions on raw feature set at timestamp $t$       & $P_{raw, bbox}^t$                       \\

                  & filtered pseudo bounding boxes at timestamp $t$       & $P_{bbox}^t$                      \\
                  & filtered pseudo hard-class labels at timestamp $t$    & $P_{cls}^t$                       \\
                  & filtered pseudo soft-class labels at timestamp $t$    & $P_{soft}^t$
                  \\ \hline
\end{tabular}}
\end{table*}

\begin{table*}[]
\centering
\caption{Implementation details on training and evaluation protocols of SSVOD. The provided values are chosen based on empirical study on ImageNet-VID dataset~\cite{ILSVRC15}.}
\label{t2}
\scalebox{0.9}{
\begin{tabular}{c|c|c}
\hline 
\multicolumn{2}{c|}{Variables}                        & Value      \\ \hline
\multirow{10}{*}{Training}  & image size in pixels (height, width)          & (1000, 600) \\
                            & number of \textit{key} image/set        & 1           \\
                            & number of \textit{reference} images/set & 2           \\
                            & \textit{reference frame} timestamp range shifted from \textit{key frame} & [-9, 9]     \\
                            & training iterations    & 40000       \\
                            & labeled set/batch      & 1           \\
                            & unlabeled set/batch    & 1           \\
                            & optimizer              & SGD         \\
                            & learning rate          & 0.005       \\
                            & EMA momentum           & 0.99        \\ \hline
\multirow{4}{*}{Evaluation} & image size in pixels (height, width)            & (1000, 600) \\
                            & number of \textit{reference} images/set & 30          \\
                            & \textit{reference frame} timestamp range shifted from \textit{key} frame  & [-15, 15]   \\
                            & model                  & student    \\ \hline
\end{tabular}}
\end{table*}

\begin{table*}[]
\centering
\caption{Summary of the data augmentations used in SSVOD training. ``-'' denotes no augmentation is applied. We apply sequential augmentation on labeled and unlabeled sets where same augmentation parameters are used for each \textit{key} and \textit{reference frame} for a particular set. We extended the standard augmentations extensively used in semi-supervised image object detection~\cite{soft, stac} to operate with videos.}
\label{t3}

\begin{tabular}{c|c|c|c}
\hline
\multirow{2}{*}{\begin{tabular}[c]{@{}c@{}}Sequential \\ Augmentation\end{tabular}} & \multirow{2}{*}{Labeled set training}
& \multirow{2}{*}{\begin{tabular}[c]{@{}c@{}}Unlabeled set training \\ (strong augmentation)\end{tabular}} & \multirow{2}{*}{\begin{tabular}[c]{@{}c@{}}Unlabeled set training\\ (weak augmentation)\end{tabular}} \\
                                                                                    &                                                                                  &                                                                                                        &                                                                                                      \\ \hline
Random flip                                                                         & $p=0.5$, ratio $\in (0, 1)$                                                           & $p=0.5$, ratio $\in (0, 1)$                                                                                 & $p=0.5$, ratio $\in (0, 1)$                                                                               \\
Contrast jitter                                                                     & $p=0.1$, ratio $\in (0, 1)$                                                           & $p=0.1$, ratio $\in (0, 1)$                                                                                 & -                                                                                                    \\
Equalize jitter                                                                     & $p=0.1$, ratio $\in (0, 1)$                                                           & $p=0.1$, ratio $\in (0, 1)$                                                                                 & -                                                                                                    \\
Solarize jitter                                                                     & $p=0.1$, ratio $\in (0, 1)$                                                           & $p=0.1$, ratio $\in (0, 1)$                                                                                 & -                                                                                                    \\
Brightness jitter                                                                   & $p=0.1$, ratio $\in (0, 1)$                                                           & $p=0.1$, ratio $\in (0, 1)$                                                                                 & -                                                                                                    \\
Sharpness jitter                                                                    & $p=0.1$, ratio $\in (0, 1)$                                                           & $p=0.1$, ratio $\in (0, 1)$                                                                                 & -                                                                                                    \\
Random posterize                                                                    & $p=0.1$, ratio $\in (0, 1)$                                                           & $p=0.1$, ratio $\in (0, 1)$                                                                                 & -                                                                                                    \\
Translation                                                                         & -                                                                                & $p=0.3$, ratio $\in (-0.1, 0.1)$                                                                            & -                                                                                                    \\
Rotation                                                                            & -                                                                                & $p=0.3$, ratio $\in (-30, 30)$                                                                              & -                                                                                                    \\
Shear                                                                               & -                                                                                & $p=0.3$, ratio $\in (-30, 30)$                                                                              & -                                                                                                    \\
Cutout                                                                              & -                                                                                & num $\in (1, 5)$, ratio $\in (0, 0.2)$                                                                       & -                                                                                                   \\ \hline
\end{tabular}
\end{table*}

\begin{table*}[]
\centering
\caption{Performance comparison with objects of different sizes and motions. Here, mAP@0.5:0.95 score is reported.}
\label{t8}
\scalebox{1.0}{
\begin{tabular}{lcccccc}
\hline
\multirow{2}{*}{Method} & \multicolumn{3}{c}{Object Size}                                 & \multicolumn{3}{c}{Motion}                                    \\ \cmidrule(lr){2-4} \cmidrule(lr){5-7}
                        & \multicolumn{1}{c}{Small} & \multicolumn{1}{c}{Middle} & Large & \multicolumn{1}{c}{Fast} & \multicolumn{1}{c}{Medium} & Slow \\ \hline
Supervised              & \multicolumn{1}{c}{6.7}  & \multicolumn{1}{c}{16.5}   & 34.5  & \multicolumn{1}{c}{13.8} & \multicolumn{1}{c}{23.9}   & 35.8 \\
Ours                    & \multicolumn{1}{c}{\textbf{8.8}}  & \multicolumn{1}{c}{\textbf{20.5}}   & \textbf{43.7}  & \multicolumn{1}{c}{\textbf{16.1}} & \multicolumn{1}{c}{\textbf{28.6}}   & \textbf{45.9} \\ \hline
\end{tabular}}
\end{table*}

\begin{table*}[]
\centering
\caption{Ablation study on the effect of different choice of thresholds for selecting pseudo  bounding box, hard, and soft-class labels.}
\label{t7}
\scalebox{1.0}{
\begin{tabular}{cccccc}
\hline
$\gamma_c$   & $\zeta_{IoU}$  & $\eta_{div}$    & mAP  & mAP@0.5 & mAP@0.75 \\ \hline
0.8 & 0.8  & 0.005 & 38.2 & 61.4    & 41.3     \\
0.8 & 0.9  & 0.005 & \textbf{39.2} & {63.8}    & \textbf{43.1}    \\
0.9 & 0.9  & 0.005 & 38.5 & 61.7    & 42.6    \\
0.8 & 0.9  & 0.01  & 38.4 & 62.0    & 42.3     \\
0.8 & 0.85 & 0.005 & 38.7 & \textbf{63.9}    & {42.4}     \\ \hline 
\end{tabular}}
\end{table*}

\begin{figure*}[t]
  \centering
   \includegraphics[width=0.9\linewidth]{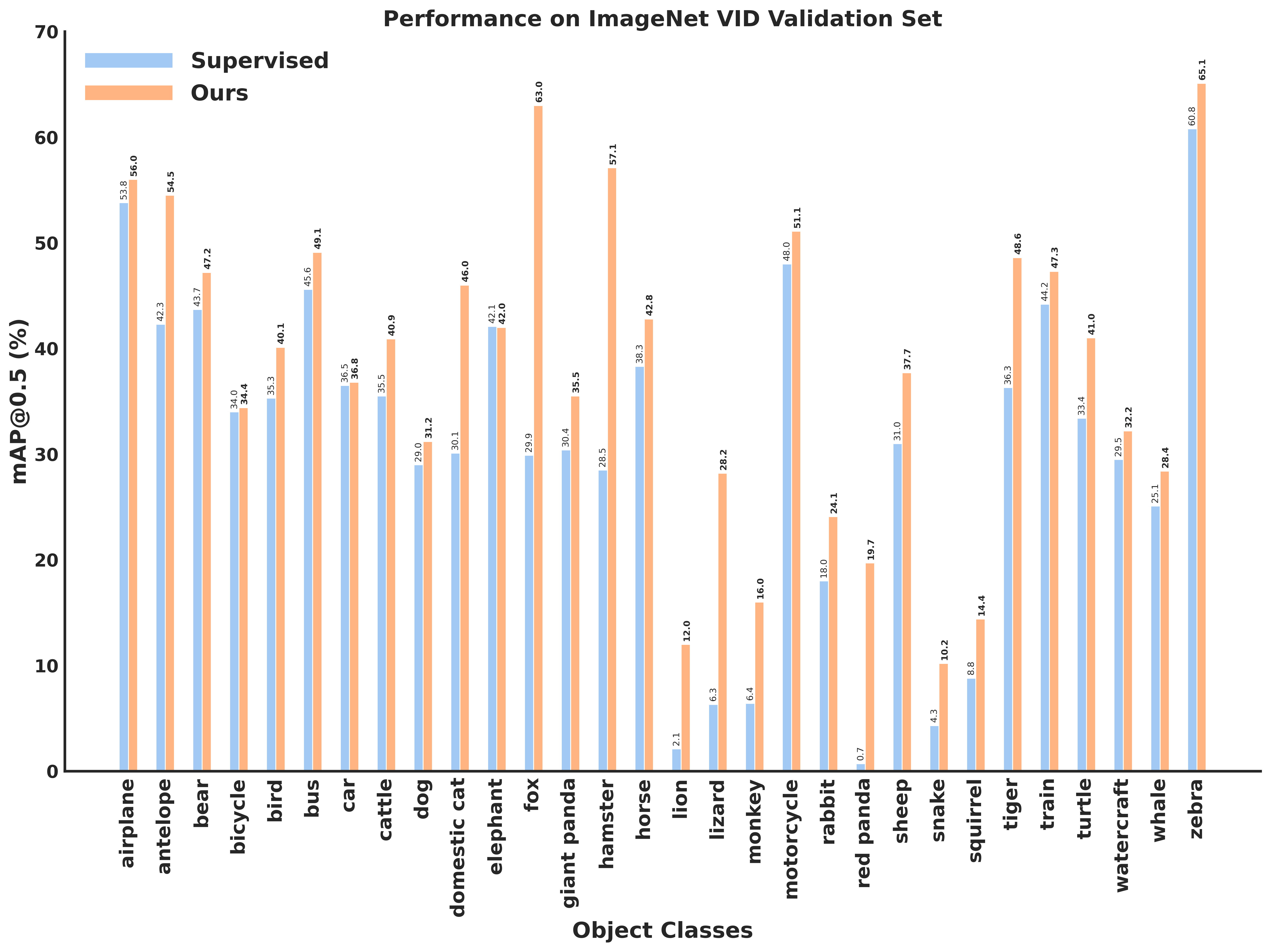}
   \caption{Per-class mAP performance comparison on the ImageNet-VID validation set~\cite{ILSVRC15} between the supervised and our proposed SSVOD approaches. In general, SSVOD significantly improves the supervised performance. Performance gain is more prominent on the challenging classes.}
   \label{f5}
\end{figure*}

\end{document}